%% file: DFV2_ICPR2026.tex
\documentclass[runningheads]{llncs}

\usepackage[T1]{fontenc}
\usepackage[utf8]{inputenc}
\usepackage{graphicx}
\usepackage{booktabs}
\usepackage{tabularx}
\usepackage{array}
\usepackage{adjustbox}
\usepackage{multirow}
\usepackage{enumitem}
\usepackage{subcaption}
\usepackage{siunitx}
\usepackage{amssymb}
\usepackage{longtable}
\usepackage{pdflscape}
\usepackage[hidelinks]{hyperref}
\usepackage{rotating}
\usepackage{multicol}
\usepackage{caption}
\usepackage{tikz}
\usetikzlibrary{arrows.meta,positioning}

\begin{document}

\title{DeepForestVisionV2: Ecology-Driven Taxonomy Expansion for Camera-Trap Monitoring in African Tropical Forests}
\titlerunning{DeepForestVisionV2: Camera-Trap Monitoring in African Tropical Forests}
\authorrunning{H. Magaldi et al.}

\author{
Hugo Magaldi\inst{1,2} \and
Th\'eau d'Audiffret\inst{1,2} \and
Etienne Fran\c{c}ois Akomo-Okoue\inst{3,4} \and
Bala Amarasekaran\inst{5} \and
Naomi Anderson\inst{5} \and
Claire Auger\inst{1,2} \and
No\'emie Cappelle\inst{6} \and
Daniel Corn\'elis\inst{7} \and
Rapha\"el Cornette\inst{8} \and
Tobias Deschner\inst{9,10} \and
Gabriel Dubus\inst{1,2} \and
Davy Fonteyn\inst{7} \and
Rosa M.\ Garriga\inst{5} \and
Jennifer Hatlauf\inst{11} \and
Innocent Kasekendi\inst{2} \and
Raymond Katumba\inst{2} \and
Aram Kazandjian\inst{5} \and
Alfred Ngomanda\inst{3} \and
St\'ephan Ntie\inst{12,13} \and
Simone Pika\inst{10} \and
Xavier Rufray\inst{6} \and
Harold Rugonge\inst{2} \and
John Justice Tibesigwa\inst{14} \and
Peter van Lunteren\inst{15} \and
Hadrien Vanthomme\inst{7} \and
Joeri A.\ Zwerts\inst{16} \and
Sabrina Krief\inst{1,2}
}

\institute{
UMR7206 Eco-Anthropologie, MNHN, Paris, France; One Forest Vision initiative \and
Sebitoli Chimpanzee Project, Sebitoli, Kibale National Park, Uganda \and
Centre National de la Recherche Scientifique et Technologique, Libreville, Gabon \and
Institut de Recherche en Ecologie Tropicale, Libreville, Gabon \and
Tacugama Chimpanzee Sanctuary, Freetown, Sierra Leone \and
International Department, Biotope, Montpellier, France \and
CIRAD, UPR For\^ets et Soci\'et\'es, Montpellier, France \and
Institut de Syst\'ematique, Evolution, Biodiversit\'e (ISYEB), UMR7205 CNRS, MNHN, SU, EPHE-PSL, UA, Paris, France \and
Max Planck Institute for Evolutionary Anthropology, Leipzig, Germany \and
Comparative BioCognition, Institute of Cognitive Science, Osnabr\"uck University, Osnabr\"uck, Germany \and
BOKU University, Institute of Wildlife Biology and Game Management, Vienna, Austria \and
D\'epartement de Biologie, Universit\'e des Sciences et Techniques de Masuku, Franceville, Gabon \and
Agence Nationale des Parcs Nationaux du Gabon, Gabon \and
Uganda Wildlife Authority, Kampala, Uganda \and
Addax Data Science, Utrecht, The Netherlands \and
Wildlife Ecology and Nature Restoration, Utrecht University, The Netherlands
}

\maketitle

\noindent{\footnotesize\textbf{Published version:} This preprint corresponds to the chapter in
\emph{Pattern Recognition. ICPR 2026 International Workshops}, LNCS 17113,
pp.~252--265 (2027). Please cite the published version:
\href{https://doi.org/10.1007/978-3-032-39518-4_17}{doi:10.1007/978-3-032-39518-4\_17}.\par}
\medskip

\begin{abstract}
Camera-trap monitoring in African tropical forests increasingly extends beyond closed-canopy interiors to riverbanks, clearings, and park edges. Among available open tools for African forest camera-trap classification, DeepForestVision is the only one providing a matched offline workflow for both photographs and videos, and previous work showed that it outperformed other available baselines on a comparable video benchmark. However, it was originally designed for closed-canopy, ground-level forest interiors and uses a 35-class prediction space that becomes too coarse when deployments encounter arboreal primates, birds, semi-aquatic taxa, or human-associated confounders such as livestock. We present DeepForestVisionV2, an ecology-driven expansion from 35 to 64 prediction classes (61 animal classes plus \emph{human}, \emph{vehicle}, and \emph{blank}) designed to address three recurrent deployment gradients: vertical stratification, scene openness, and anthropogenic interfaces. DeepForestVisionV2 retains the same offline workflow and is trained on 1,535,010 photographs and 243,354 videos from multi-country African tropical-forest projects. Evaluation combines a cross-country cropped-photo validation set, used to assess robustness across sites and camera-trap settings, with three held-out Uganda video benchmarks spanning the targeted gradients. On the validation set, DeepForestVisionV2 reaches 0.86 accuracy, 0.82 macro-F1, and 0.81 balanced accuracy. On the deployment benchmarks, it preserves or improves baseline accuracy despite its harder classification task, while increasing the number of identified taxa from 22 to 29 in forest-interior videos and from 4 to 9 at riverbanks. In the park-edge use case, it raises accuracy from 0.62 to 0.86 and reduces false alarms from 11 to 0. These results show that DeepForestVisionV2 materially improves field utility while preserving robustness across sites, habitats, and camera-trap settings.

\end{abstract}

\section{Introduction}

Camera-traps are now a standard instrument for wildlife monitoring because they support species inventories, activity analyses, and disturbance assessment at spatial and temporal scales that are difficult to achieve by direct observation \cite{tropical_forests_monitoring}. As the volume of collected data grows, automated species identification becomes essential for practical workflows \cite{PlatformsEvaluation}. In practice, the ecological value of identification tools depends on more than their accuracy: a system must be deployable under field constraints, remain robust across the habitats where camera-traps are actually placed, and provide the taxonomic resolution required by downstream ecological questions \cite{tuia_perspectives_2022,whytock2021robust}. 

For African tropical forests, several open tools are available, including Mbaza \cite{whytock2021robust}, Zamba \cite{dorne2025zamba}, SpeciesNet \cite{SpeciesNet}, and DeepForestVision \cite{DFV}. These systems differ in taxonomy, but also in deployment constraints: Mbaza and SpeciesNet process photographs only, whereas Zamba requires online or code-based local execution. DeepForestVision is the only tool for non-programmers that provides offline processing of both photographs and videos through AddaxAI, a no-code interface for local model deployment \cite{vanLunteren2023,DFV}. Previous work already compared DeepForestVision with Mbaza, Zamba, and SpeciesNet on a video benchmark from Kibale National Park, Uganda, by adding the engineering needed to run photo-native methods on videos, and found that DeepForestVision outperformed all three in accuracy (Mbaza by 45.0\%, Zamba by 13.1\% and SpeciesNet by 37.7\%) \cite{DFV}. Accordingly, this paper does not aim to establish another cross-tool benchmark. Instead, it shows that an ecology-driven expansion of the DeepForestVision label space improves field utility within the same offline workflow for photographs and videos.

DeepForestVision was designed around closed-canopy ground-level deployments. However, ongoing projects increasingly deploy camera-traps along ecological gradients, understood here as recurrent changes in deployment conditions that shift the observed community composition: (i) a \textit{vertical gradient}, where arboreal primates become visible near clearings, mineral licks, or canopy openings \cite{Moore_arboreal,pebsworth_selecting_2021}; (ii) an \textit{openness gradient}, where riverbanks and open views increase detections of birds \cite{obrien2008birds} and expose cameras to riparian and semi-aquatic taxa; and (iii) an \textit{anthropogenic gradient}, where park-edge cameras must distinguish wildlife from livestock \cite{mckaughan2025cropprimates,walton_camera_2022,whytock2023realtime}. In these regimes, DeepForestVision's coarse labels such as \emph{bird}, \emph{monkey}, or \emph{civet/genet} are often insufficient for ecological interpretation or management action. DeepForestVisionV2 addresses this mismatch by redesigning the label space to better capture taxa that become relevant along these deployment gradients, while preserving the robustness of a field-ready workflow.

\paragraph{Contributions.} This paper makes three contributions:
\begin{enumerate}[leftmargin=1.1em]
    \item an ecology-driven expansion from 35 to 64 prediction classes (61 animal classes, plus \emph{human}, \emph{vehicle}, and \emph{blank}) aligned with three ecologically motivated deployment gradients;
    \item a geographically diverse, Africa-wide cropped-photo validation benchmark used to assess robustness of the expanded classifier across sites and camera-trap settings;
    \item a deployment-centric evaluation on three held-out Uganda video benchmarks, comparing DeepForestVisionV2 with the original DeepForestVision under a matched offline workflow.
\end{enumerate}

\section{Material and methods}

\subsection{Study system and datasets}

We assembled a multi-country corpus of labeled camera-trap data from African tropical forests intended to support geographic robustness, by combining partner projects and publicly available resources (Table~\ref{tab:data_sources}). In total, the training and validation pool contains 1,535,010 photographs and 243,354 videos acquired across different sites, habitats, camera-trap protocols, and imaging conditions. The data were collected non-invasively and required no contact with animals. All necessary authorizations and permits were obtained prior to data collection and are available upon request.

\begin{table}[t]
\caption{Training and validation data sources.}
\label{tab:data_sources}
\centering
\small
\begin{tabularx}{\textwidth}{>{\raggedright\arraybackslash}p{5.0cm} >{\raggedleft\arraybackslash}p{2.1cm} >{\raggedleft\arraybackslash}p{1cm} >{\raggedleft\arraybackslash}p{2cm} >{\raggedright\arraybackslash}X}
\toprule
\textbf{Dataset} & \textbf{Region} & \textbf{Sites} & \textbf{Type} & \textbf{Items} \\
\midrule
Sebitoli Chimpanzee Project & Uganda & 1 & Video & 165,942 \\
CIRAD-SWM \cite{CIRAD} & Gabon & 1 & Video & 24,430 \\
Biotope Baseline Biodiversit\'e & Gabon & 1 & Video & 17,046 \\
Tacugama Chimpanzee Sanctuary & Sierra Leone & 1 & Video & 15,194 \\
Ozouga Chimpanzee Project & Gabon & 1 & Video & 14,703 \\
J. A. Zwerts \cite{zwerts_fsc-certified_2024} & Congo, Gabon & 14 & Photo & 643,660 \\
Wildlife Insights \cite{ahumada2020wildlife} & Global & 10 & Photo & 617,882 \\
LLA BC & Global & 3 & Photo & 267,778 \\
Pan African Programme & Africa & 29 & Video & 6,039 \\
InterMuc Project & Global & 1 & Photo & 5,094 \\
Golden Jackal Project \cite{hatlauf_hacklaender_2016,suss_hatlauf_2024} & Global & 1 & Photo & 596 \\
\midrule
\textbf{Total photographs} &  &  &  & \textbf{1,535,010} \\
\textbf{Total videos} &  &  &  & \textbf{243,354} \\
\bottomrule
\end{tabularx}
\end{table}

Evaluation is split into two complementary parts. First, the data sources listed in Table~\ref{tab:data_sources} were transformed into cropped animal detections using MegaDetector v5 \cite{beery2019pipeline} and split into training and validation sets using a random partition at camera level prior to crop extraction, so as to reduce spurious correlations between the two sets. To limit long-tail imbalance in evaluation while preserving rare taxa for training, each class contributes at most 1{,}000 validation crops and no more than 10\% of the available crops for that class. This cropped-detection validation set is designed to measure the robustness of the classification step across countries, sites, habitats, and camera-trap settings and covers the full 61-animal target label space. 

\begin{figure}[t]
    \centering
    \begin{subfigure}[t]{0.49\textwidth}
        \centering
        \includegraphics[width=\linewidth]{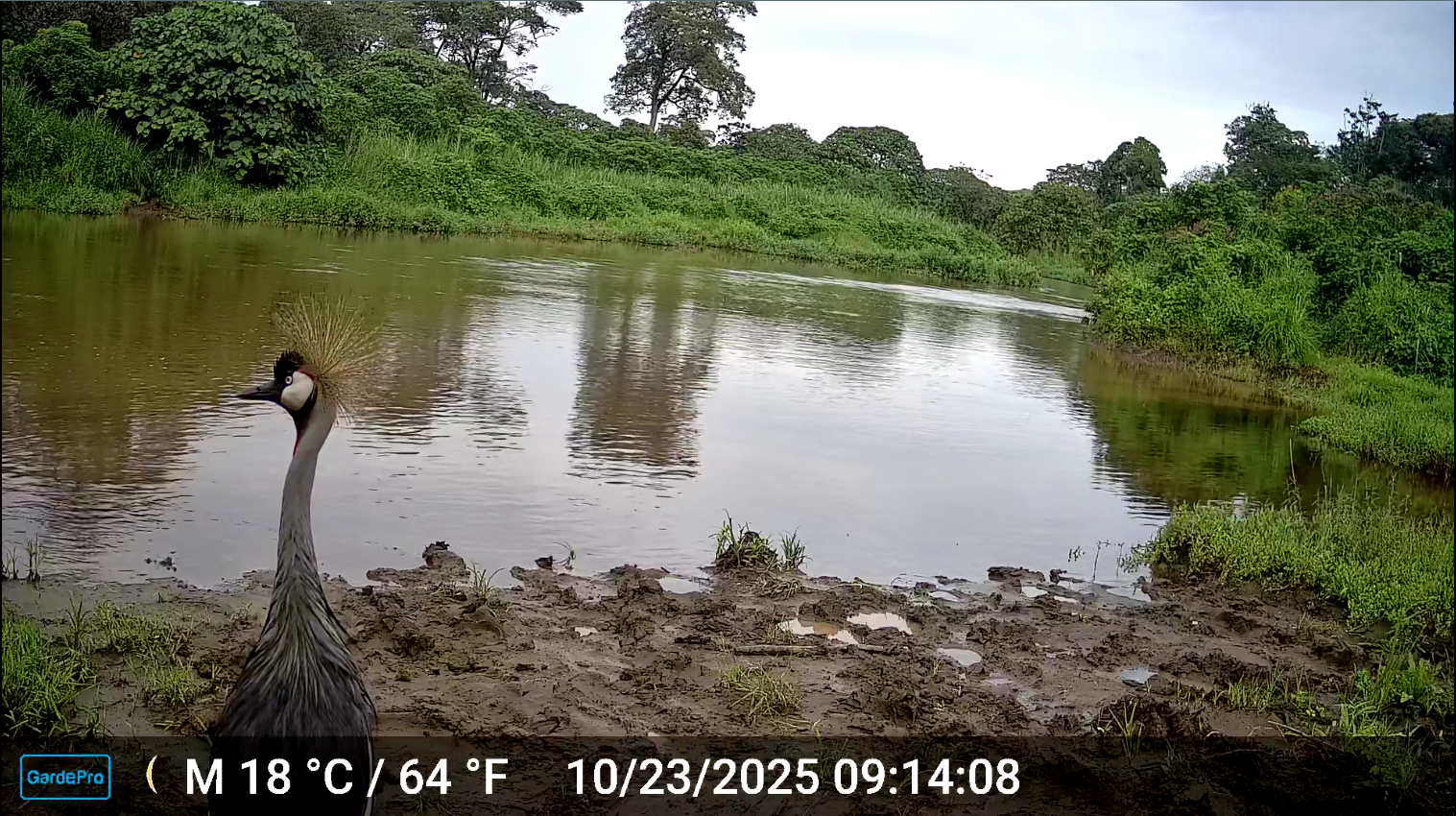}
        \caption{Riverbank view with a Grey crowned crane (\emph{Balearica regulorum}).}
    \end{subfigure}\hfill
    \begin{subfigure}[t]{0.49\textwidth}
        \centering
        \includegraphics[width=\linewidth]{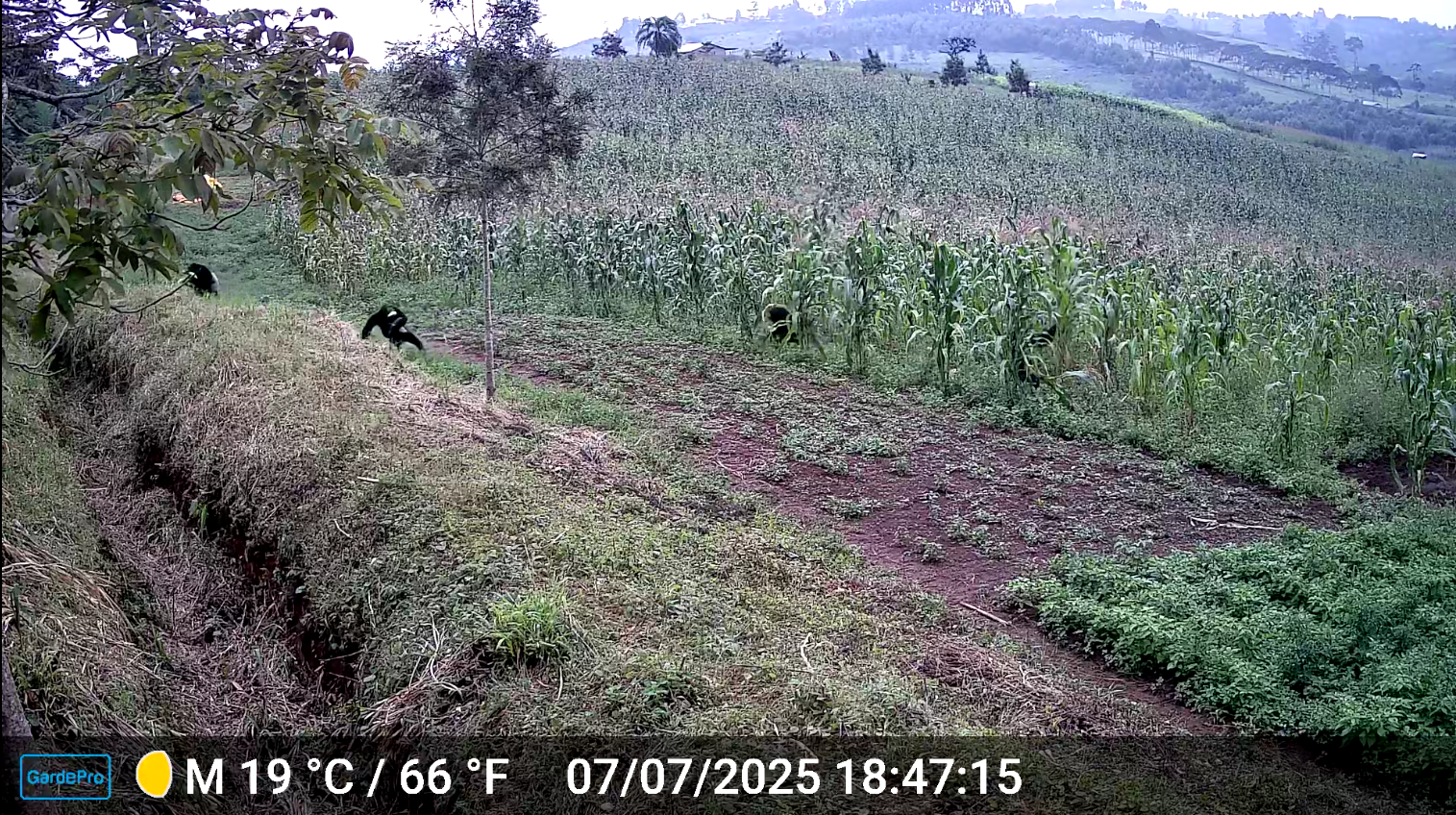}
        \caption{Chimpanzees (\emph{Pan troglodytes}) crop-feeding in maize fields.}
    \end{subfigure}
    \caption{Two deployment regimes targeted by the taxonomy expansion: openness on riverbanks and anthropogenic interfaces at park edges. \copyright\ SCP}
    \label{fig:deployment_examples}
\end{figure}

Second, to assess operational value in realistic field settings, we evaluate the whole DeepForestVisionV2 pipeline on three video benchmarks corresponding  to the three targeted ecological gradients (Table~\ref{tab:gradients_benchmarks}, Figure \ref{fig:deployment_examples}). These benchmarks were collected and annotated by the Sebitoli Chimpanzee Project in Kibale National Park, Uganda, and were not included in the training or validation pool.

\begin{table}[t]
\caption{Ecological gradients motivating the taxonomy redesign and the corresponding held-out deployment benchmarks.}
\label{tab:gradients_benchmarks}
\centering
\small
\begin{tabularx}{\textwidth}{>{\raggedright\arraybackslash}p{2.1cm} >{\raggedright\arraybackslash}X >{\raggedright\arraybackslash}p{4.7cm}}
\toprule
\textbf{Gradient} & \textbf{Ecological motivation} & \textbf{Benchmark} \\
\midrule
Vertical stratification & Arboreal primates become visible near clearings, mineral licks, and canopy openings. & Forest interior, 17,899 videos (Feb.-Jul. 2025), 29 animal classes including 9 primates. \\
Openness & Riverbanks and open views expose birds and semi-aquatic taxa that rarely appear in closed-canopy interiors. & Riverbanks, 1,016 videos (Oct.-Nov. 2025), 9 animal classes including 6 birds or semi-aquatic taxa. \\
Anthropogenic interface & Park-edge monitoring must distinguish genuine wildlife intrusions from domestic passages.  & Park-edge, 146 videos (Jul.-Sep. 2025), 3 animal classes including one domestic animal. \\
\bottomrule
\end{tabularx}
\end{table}

\subsection{Taxonomy of DeepForestVisionV2}

DeepForestVisionV2 outputs one label among 64 prediction classes: 61 animal classes spanning mammals and birds, plus \emph{human}, \emph{vehicle}, and \emph{blank}. To support finer-grained downstream ecological analyses, this redesign increases DeepForestVision's native 35-class resolution for arboreal primates, birds, riparian taxa, and domestic species (Table \ref{tab:taxonomy_expansion_groups}). The full taxonomy, including scientific names, taxonomic levels, IUCN status, training image counts, and mappings to the original DeepForestVision classes, is reported in Appendix~\ref{app:taxonomy}.

\begin{table}[t]
\caption{DeepForestVisionV2 animal classes not represented as such in DeepForestVision, grouped by the main ecological contexts. Boldface indicates newly introduced classes, non boldface refinements of previously coarser classes.}
\label{tab:taxonomy_expansion_groups}
\centering
\small
\setlength{\tabcolsep}{6pt}
\begin{tabularx}{\textwidth}{>{\raggedright\arraybackslash}p{3.2cm} X}
\toprule
\textbf{Ecological grouping} & \textbf{DeepForestVisionV2 classes} \\
\midrule
Primates &
agile mangabey; grey-cheeked mangabey; moustached monkey; red-capped mangabey; red-tailed monkey; spot-nosed monkey; red colobus; galago; potto \\
\midrule
Birds &
bird of prey; black guineafowl; casqued hornbill; crane; dove; duck; francolin; hornbill; nkulengu rail; passeriform; pelecaniform; rockfowl; turaco \\
\midrule
Semi-riparian or aquatic taxa &
\textbf{hippopotamus}; \textbf{otter}; \textbf{sitatunga} \\
\midrule
Other mammals &
black-legged mongoose; civet; genet; warthog; \textbf{cattle}; \textbf{dog}; \textbf{goat} \\
\bottomrule
\end{tabularx}
\end{table}

\subsection{Photo-and-video pipeline and training}

\paragraph{Pipeline.} DeepForestVisionV2 retains the original DeepForestVision architecture and produces one label per camera-trap item (defined as one photograph or one video). Videos are important because they are a standard output of many camera-trap deployments and field teams typically review full clips rather than isolated frames.

For each photograph, or for each sampled frame extracted from a video, MegaDetector v5 \cite{beery2019pipeline} is first applied to detect \emph{animal}, \emph{human}, and \emph{vehicle} instances. Human and vehicle detections are assigned directly from detector outputs, while animal detections are cropped and sent to a DINOv3 ViT-B/16 classifier \cite{simeoni2025dinov3}. Class scores are computed on retained detections across sampled frames and then averaged to produce one item-level prediction. If no detection remains after thresholding, the item is labeled \emph{blank}.

\medskip

\noindent\emph{Training.} The detector stage was used without additional training. For classification, we fine-tuned a pretrained DINOv3 ViT-B/16 on MegaDetector crops (threshold 0.5) in PyTorch. Training augmentation comprised random resized crop to $224\times224$ (scale $[0.9,1.0]$, aspect ratio $[0.9,1.1]$), random horizontal flip ($p=0.5$), color jitter ($p=0.2$; brightness $\pm0.15$, contrast $\pm0.20$, saturation $\pm0.10$), random grayscale ($p=0.05$), random autocontrast ($p=0.2$), random sharpness adjustment (factor 1.5, $p=0.2$), and Gaussian blur (kernel size 9, $\sigma\in[0.1,1.0]$, $p=0.05$). Images were converted to tensors and normalized with ImageNet statistics \cite{deng2009imagenet}; validation preprocessing used resize to 256, center crop to $224\times224$, and the same normalization. Optimization used AdamW with separate learning rates for backbone ($10^{-5}$) and head ($10^{-4}$), weight decay 0.05, label smoothing 0.05, cosine scheduling with 5\% warm-up, automatic mixed precision, and gradient clipping at 1.0. Training ran for 20 epochs, with checkpoint selection by validation balanced accuracy.

\subsection{Baseline, comparison protocol, and metrics}

\paragraph{Baseline.}
The original DeepForestVision is the baseline on the deployment benchmarks because it is the only available open tool sharing with DeepForestVisionV2 a matched offline workflow for processing both photographs and videos.

\paragraph{Protocol.}
We evaluate the original DeepForestVision in its native 35-class label space using the mapping given in Appendix~\ref{app:taxonomy}, which defines an easier classification task than the 64-class label space used to evaluate DeepForestVisionV2. The six DeepForestVisionV2 classes with no corresponding DeepForestVision class -\emph{cattle}, \emph{dog}, \emph{goat}, \emph{hippopotamus}, \emph{otter}, and \emph{sitatunga} (Table~\ref{tab:classes})- are retained, because the goal is to quantify the ecological utility gained from the expanded DeepForestVisionV2 label space.

\paragraph{Metrics.}  We report accuracy, balanced accuracy, and macro-averaged F1-score as global metrics, along with per-class precision, recall, F1-score, and support. Since the DeepForestVisionV2 class distribution is strongly long-tailed (Appendix~\ref{app:taxonomy}), balanced accuracy and macro-F1 are given to reflect performance beyond the most frequent taxa. Blank and human items are excluded so that the reported scores reflect differences in animal classification rather than shared detector outputs, since both labels are assigned by MegaDetector and are therefore identical for DeepForestVision and DeepForestVisionV2.

\paragraph{Ecological utility.} To quantify ecological gain, we also report the number and identity of taxa surfaced by DeepForestVisionV2 but not represented as such in DeepForestVision and benchmark performance across the three motivating ecological gradients. For the park-edge benchmark containing primates and goats, we design a simple alarm task in which the alarm group contains all primates, and goat passages are nuisance confounders because they trigger unnecessary interventions if the system cannot separate them from wildlife. We report true alarms, false alarms, and missed alarms for primate intrusions, as well as precision and recall computed from the alarm counts. 

\paragraph{Calibration.} Confidence calibration is assessed with a reliability diagram and the Expected Calibration Error (ECE), which measures the average absolute difference between predicted confidence and empirical accuracy across confidence bins. Since model scores may be used for review prioritization, filtering, or semi-automated alerting in field deployment, calibration is operationally important.

\section{Results}

\subsection{Validation set.}

On the cross-country validation set, used here to assess robustness of the expanded classifier over the full animal label space, DeepForestVisionV2 reaches 0.86 accuracy, 0.81 balanced accuracy, and 0.82 macro-F1. All per-class validation supports and metrics are reported in Appendix~\ref{app:classmetrics}.

\subsection{Deployment benchmarks and surfaced taxa}

Global performance metrics on the deployment benchmarks are summarized in Table~\ref{tab:summary_metrics}. On the forest-interior benchmark, DeepForestVisionV2 matches the original DeepForestVision in accuracy (0.89 for both models) while increasing the number of surfaced taxa from 22 to 29. On riverbanks, DeepForestVisionV2 remains close in accuracy (0.72 vs. 0.73) and raises the number of surfaced taxa from 4 to 9. On the park-edge benchmark, DeepForestVisionV2 improves accuracy from 0.62 to 0.86.

\begin{table}[t]
\caption{Deployment-benchmark metrics for DeepForestVisionV2 (64 classes) and DeepForestVision (35 classes).}
\label{tab:summary_metrics}
\centering
\small
\begin{tabularx}{\textwidth}{>{\raggedright\arraybackslash}p{2.3cm} >{\raggedright\arraybackslash}p{2.2cm} >{\raggedleft\arraybackslash}X >{\raggedleft\arraybackslash}X >{\raggedleft\arraybackslash}X}
\toprule
\textbf{Benchmark} & \textbf{Model} & \textbf{Acc.} & \textbf{Bal. acc.} & \textbf{Macro-F1} \\
\midrule
\multirow{2}{*}{Forest interior}
& DeepForestVisionV2 & \textbf{0.89} & \textbf{0.77} & \textbf{0.73} \\
& DeepForestVision & \textbf{0.89} & 0.70 & 0.70 \\
\midrule
\multirow{2}{*}{Riverbanks}
& DeepForestVisionV2 & 0.72 & \textbf{0.71} & \textbf{0.59} \\
& DeepForestVision & \textbf{0.73} & 0.63 & 0.55 \\
\midrule
\multirow{2}{*}{Park-edge}
& DeepForestVisionV2 & \textbf{0.86} & \textbf{0.86} & \textbf{0.92} \\
& DeepForestVision & 0.62 & 0.62 & 0.62 \\
\bottomrule
\end{tabularx}
\end{table}

\begin{table}[t]
\caption{Taxa surfaced by DeepForestVisionV2 that are absent as such from the DeepForestVision label space, with weighted F1 per ecological group.}
\label{tab:new_taxa}
\centering
\small
\begin{tabularx}{\textwidth}{>{\raggedright\arraybackslash}p{3cm} >{\raggedright\arraybackslash}X >{\raggedleft\arraybackslash}p{2.2cm} >{\raggedleft\arraybackslash}p{1.0cm}}
\toprule
\textbf{Group} & \textbf{Taxa} & \textbf{Weighted F1} & \textbf{Support} \\
\midrule
Interior - primates & galago, grey-cheeked mangabey, red colobus, red-tailed monkey & 0.78 & 265 \\
Interior - birds & bird of prey, dove, francolin & 0.79 & 100 \\
Interior - other & civet, genet, hippopotamus, dog & 0.78 & 105 \\
\midrule
Riverbanks - birds & bird of prey, crane, duck, pelecaniform & 0.53 & 20 \\
Riverbanks - other & hippopotamus, otter & 0.44 & 15 \\
\midrule
Park-edge - domestic & goat & 0.86 & 50 \\
\bottomrule
\end{tabularx}
\end{table}

Table~\ref{tab:new_taxa} details the additional taxa recovered by the expanded label space on the video benchmarks. On the forest-interior benchmark, these added detections include several arboreal primates and bird groups that were previously merged into coarser labels. On riverbanks, the additional detections are concentrated in bird and semi-aquatic taxa associated with open riparian views.

\subsection{Operational use case: filtering crop-raiding alarms}

\begin{table}[t]
\caption{Alarm-filtering performance on the park-edge benchmark. Alarm precision and recall are derived from the true, false, and missed alarm counts.}
\label{tab:alarms}
\centering
\small
\begin{tabularx}{\textwidth}{>{\raggedright\arraybackslash}p{3.0cm} >{\raggedleft\arraybackslash}p{1.5cm} >{\raggedleft\arraybackslash}p{1.5cm} >{\raggedleft\arraybackslash}p{1.7cm} >{\raggedleft\arraybackslash}p{1.3cm} >{\raggedleft\arraybackslash}X}
\toprule
\textbf{Model} & \textbf{True alarms} & \textbf{False alarms} & \textbf{Missed alarms} & \textbf{Precision} & \textbf{Recall} \\
\midrule
DeepForestVisionV2 & 91 & 0 & 5 & 1.00 & 0.95 \\
DeepForestVision & 93 & 11 & 3 & 0.89 & 0.97 \\
\bottomrule
\end{tabularx}
\end{table}

On the park-edge benchmark, DeepForestVisionV2 removes all false alarms while keeping high recall, and increases missed alarms from 3 to 5 (Table~\ref{tab:alarms}). Overall, it raises benchmark accuracy from 0.62 to 0.86 (Table~\ref{tab:summary_metrics}).

\subsection{Calibration}

\begin{figure}[h]
    \centering
    \includegraphics[width=0.5\textwidth]{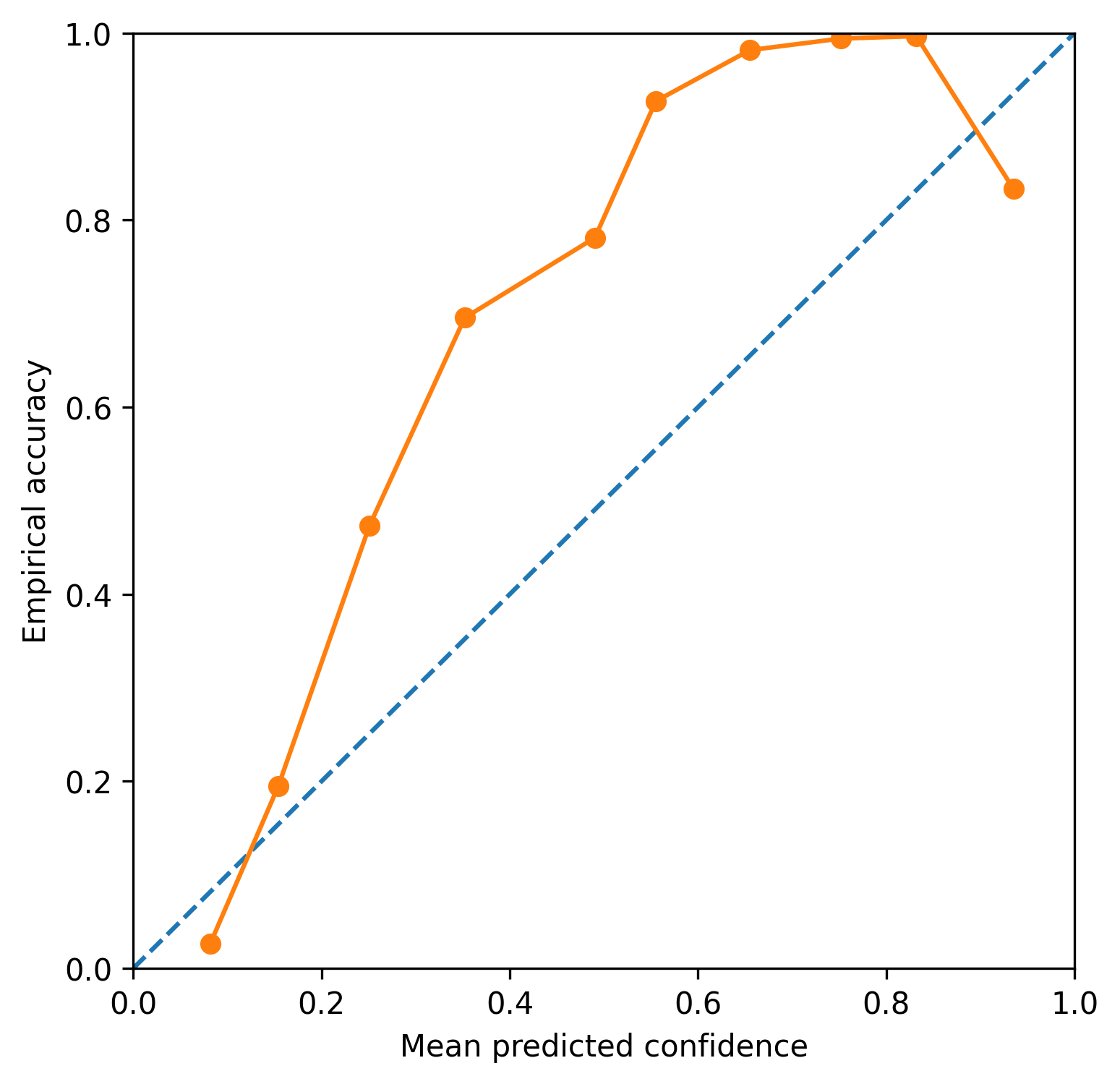}
    \caption{Reliability diagram for pooled benchmark predictions: empirical accuracy by confidence bin (orange) versus perfect calibration (dashed blue).}
    \label{fig:reliability}
\end{figure}

On pooled video-benchmark predictions, DeepForestVisionV2 has an Expected Calibration Error of 0.17. The reliability diagram in Figure~\ref{fig:reliability} shows that, over most confidence bins, empirical accuracy is higher than predicted confidence, indicating under-confidence rather than over-confidence. The apparent drop in the highest-confidence bin is unstable because it contains only six predictions.

\section{Discussion}\label{sec:discussion}

On the geographically diverse validation set, DeepForestVisionV2 remains strong in the full animal target space, which indicates that the expanded taxonomy classification is robust across sites, habitats and camera-trap settings. However, classes with low training support are more likely to exhibit low F1 scores (worst performers are \emph{honey badger} 0.42, \emph{potto} 0.49 and \emph{red-tailed monkey} 0.58) and would benefit from further training on additional data. Performance on the video benchmarks is lower than cropped-image classification performance on the validation set, showing that validation metrics are informative but optimistic with respect to end-to-end deployment. This difference is expected in part because the video pipeline also depends on detection quality, and therefore includes errors from missed or incorrect detections in addition to classification errors. It may be further amplified by limited validation support for some taxa, with nine classes represented by fewer than 100 validation images, and by calibration choices made on the validation set. Validation performance is therefore best interpreted as a controlled comparison signal rather than a direct estimate of global field performance.

Our video benchmarks ask an operational workflow question under field conditions in Uganda. Across the three deployment gradients, DeepForestVisionV2 preserves or increases the accuracy of DeepForestVision despite its harder classification task, and provides different kinds of ecological gains. In the forest interior, DeepForestVisionV2 surfaces additional arboreal primates and birds. On riverbanks, the benefit is primarily taxonomic coverage in an open-scene regime where the original taxonomy was not designed to operate. However, this benchmark is relatively small and should be expanded to additional cameras, seasons, and riparian settings. At the park edge, the benefit is directly operational: domestic classes suppress nuisance alarms and make the system more suitable for future edge-AI alerting pipelines in which review capacity is limited and false positives are costly \cite{mckaughan2025cropprimates,whytock2023realtime}. DeepForestVisionV2 removes all 11 false alarms triggered by DeepForestVision, but adds 2 more missed alarms. This trade-off suggests that calibration and threshold tuning are practical steps needed before fully automated alerting.

The reliability diagram and Expected Calibration Error show that DeepForestVisionV2 is underconfident on the deployment benchmarks, and suggest that confidence scores should not be interpreted as calibrated probabilities. Post-hoc calibration, such as temperature scaling on held-out data, would be advisable before using confidence thresholds for automated triage or alarming. 
Finally, the pipeline as we present it here emits a single label per clip, which is acceptable for most present deployments but will miss the occasional multi-species event. This can be changed in the AddaxAI interface to output all predictions above a set confidence threshold. 

\section{Conclusion}

DeepForestVisionV2 expands DeepForestVision from a 35-class forest-interior camera-trap model to a 64-class system better aligned with field realities in African tropical forests. This expansion is motivated by three recurrent ecological gradients - vertical stratification, scene openness, and anthropogenic interfaces - and evaluated on benchmarks matched to each of them. On a cross-country validation set, DeepForestVisionV2 reaches 0.86 accuracy, 0.81 balanced accuracy, and 0.82 macro-F1 over the full animal target space. On held-out video benchmarks, it preserves or improves performance despite the harder classification task, while surfacing more taxa in forest-interior and riverbank settings and providing a direct operational benefit for park-edge alarm filtering. Together, these results indicate that ecology-driven taxonomy expansion increases the practical value of a field-ready offline camera-trap workflow and makes it more informative for conservation use across contrasting habitats.

\paragraph{Data availability.} Camera-trap imagery cannot be released publicly because of data-sharing agreements and permitting restrictions. Derived, non-identifying outputs such as label mappings and evaluation scripts are available from the authors on reasonable request. DeepForestVisionV2 is openly available through the offline no-code interface AddaxAI \url{https://addaxdatascience.com/addaxai}. Model weights and inference code are provided through the OneForestVision initiative GitHub repository: \url{https://github.com/MNHN-OFVI/DeepForestVisionV2}. 

\paragraph{Acknowledgements.} The authors thank the Fondation pour la Nature et l'Homme, Fondation Prince Albert 2 de Monaco, Fonds Français pour l'Environnement Mondial, and the One Forest Vision initiative, funded by the French Ministry of Higher Education and Scientific Research and the Ministry of Europe and Foreign Affairs, for supporting this work. Tacugama Chimpanzee Sanctuary was supported by USFS, USFWS, and USAID, with the collaboration of the National Protected Area Authority of Sierra Leone. In Gabon, the authors thank ANPN, CENAREST, PWG-CEB, Station d'Etudes des Gorilles et Chimpanzés, and Lopé National Park for authorizations and access to the study sites, and acknowledge support from the European Union through the Sustainable Wildlife Management Programme, the Wild Chimpanzee Foundation, the Sievert Foundation, the Commonwealth Scholarship Commission UK, the University of Oxford's Hertford College Mortimer May Senior Scholarship in Geography, D. Lehmann, J. Edzang Ndong, E. Dimoto, and the Zooniverse team and citizen scientists.

\bibliographystyle{splncs04}
\bibliography{DFV2}

\newpage
\appendix
\input{appendices}

\end{document}

%% file: appendices.tex
\begin{landscape}
\section{Full class taxonomy and mapping}\label{app:taxonomy}

\scriptsize
\setlength{\LTleft}{0pt}
\setlength{\LTright}{0pt}
\begin{longtable}{@{}p{3.9cm}p{3.2cm}p{2.8cm}p{0.8cm}r p{4cm}p{1.5cm}@{}}
\caption{DeepForestVisionV2 class taxonomy with IUCN status, training support, and mapping to DeepForestVision. Classes with fewer than 1{,}000 training images are marked with $^\dagger$. IUCN codes: LC, NT, VU, EN, CR; NA = domestic taxa not assessed by IUCN. For higher taxa, $\geqslant X$ denotes the least threatened member species ranging in African tropical forests.}\label{tab:classes}\\
\toprule
\textbf{Class DeepForestVisionV2} & \textbf{Scientific name} & \textbf{Taxonomic level} & \textbf{IUCN} & \textbf{Support} & \textbf{Class DeepForestVision} & \textbf{Mapping} \\
\midrule
\endfirsthead
\multicolumn{7}{@{}l}{\tablename~\thetable{} (continued)}\\
\toprule
\textbf{Class DeepForestVisionV2} & \textbf{Scientific name} & \textbf{Taxonomic level} & \textbf{IUCN} & \textbf{\# Training Images} & \textbf{Class DeepForestVision} & \textbf{Mapping} \\
\midrule
\endhead
\bottomrule
\endfoot
\bottomrule
\endlastfoot
aardvark & \emph{Orycteropus afer} & species & LC & 1,328 & aardvark &  \\
african buffalo & \emph{Syncerus caffer} & species & NT & 207,162 & african buffalo &  \\
african elephant & \emph{Loxodonta} sp. & genus & $\geqslant$ EN & 642,161 & african elephant &  \\
african golden cat & \emph{Caracal aurata} & species & VU & 9,816 & african golden cat &  \\
agile mangabey & \emph{Cercocebus agilis} & species & LC & 22,951 & monkey & refinement \\
baboon & \emph{Papio} sp. & genus & $\geqslant$ LC & 434,969 & baboon &  \\
bird of prey & \emph{Accipitriformes} & order & $\geqslant$ LC & 2,448 & bird & refinement \\
black guineafowl & \emph{Agelastes niger} & species & LC & 35,626 & guineafowl & refinement \\
black-and-white colobus & \emph{Colobus} sp. & genus & $\geqslant$ LC & 7,220 & black-and-white colobus &  \\
black-legged mongoose & \emph{Bdeogale nigripes} & species & LC & 7,211 & mongoose & refinement \\
blue duiker & \emph{Philantomba} sp. & genus & $\geqslant$ LC & 489,020 & blue duiker &  \\
blue monkey & \emph{Cercopithecus mitis} & species & LC & 3,590 & blue monkey &  \\
bushbuck & \emph{Tragelaphus scriptus} & species & LC & 204,197 & bushbuck &  \\
bushpig & \emph{Potamochoerus} sp. & genus & $\geqslant$ LC & 196,209 & bushpig &  \\
casqued hornbill & \emph{Ceratogymna atrata} & species & NT & 1,156 & bird & refinement \\
cattle & \emph{Bos taurus} & species (domestic) & NA & 86,688 &  & new class \\
chimpanzee & \emph{Pan troglodytes} & species & EN & 198,350 & chimpanzee &  \\
civet & \emph{Civettictis civetta} & species & LC & 2,784 & civet\_genet & refinement \\
crane$^\dagger$ & \emph{Gruidae} & family & $\geqslant$ LC & 921 & bird & refinement \\
dog & \emph{Canis lupus familiaris} & subspecies (domestic) & NA & 16,624 &  & new class \\
dove & \emph{Columbidae} & family & $\geqslant$ LC & 8,012 & bird & refinement \\
duck & \emph{Anatidae} & family & $\geqslant$ LC & 3,205 & bird & refinement \\
francolin & \emph{Pternistis} & genus & $\geqslant$ LC & 9,379 & bird & refinement \\
galago & \emph{Galagidae} & family & $\geqslant$ LC & 1,512 & galago\_potto & refinement \\
genet & \emph{Genetta} & genus & $\geqslant$ LC & 16,014 & civet\_genet & refinement \\
goat & \emph{Capra hircus} & species (domestic) & NA & 21,358 &  & new class \\
gorilla & \emph{Gorilla} sp. & genus & $\geqslant$ CR & 24,099 & gorilla &  \\
grey-cheeked mangabey & \emph{Lophocebus albigena} & species & LC & 5,698 & monkey & refinement \\
guineafowl & \emph{Numididae} & family & $\geqslant$ LC & 181,440 & guineafowl &  \\
hippopotamus & \emph{Hippopotamus amphibius} & species & VU & 6,817 &  & new class \\
honey badger & \emph{Mellivora capensis} & species & LC & 1,959 & honey badger &  \\
hornbill$^\dagger$ & \emph{Bucerotidae} & family & $\geqslant$ LC & 963 & bird & refinement \\
hyrax & \emph{Procaviidae} & family & $\geqslant$ LC & 8,090 & hyrax &  \\
leopard & \emph{Panthera pardus} & species & VU & 7,134 & leopard &  \\
lhoests monkey & \emph{Allochrocebus lhoesti} & species & VU & 30,919 & lhoests monkey &  \\
mandrill & \emph{Mandrillus sphinx} & species & VU & 82,529 & mandrill &  \\
mongoose & \emph{Herpestidae} & family & $\geqslant$ LC & 23,793 & mongoose &  \\
moustached monkey & \emph{Cercopithecus cephus} & species & LC & 1,487 & monkey & refinement \\
nkulengu rail & \emph{Himantornis haematopus} & species & LC & 2,787 & bird & refinement \\
otter & \emph{Lutrinae} & subfamily & $\geqslant$ NT & 1,257 &  & new class \\
pangolin & \emph{Manidae} & family & $\geqslant$ EN & 6,832 & pangolin &  \\
passeriform & \emph{Passeriformes} & order & $\geqslant$ LC & 6,887 & bird & refinement \\
pelecaniform & \emph{Pelecaniformes} & order & $\geqslant$ LC & 4,216 & bird & refinement \\
porcupine & \emph{Hystricidae} & family & $\geqslant$ LC & 51,320 & porcupine &  \\
potto$^\dagger$ & \emph{Perodicticus potto} & species & NT & 162 & galago\_potto & refinement \\
red colobus & \emph{Piliocolobus} sp. & genus & $\geqslant$ VU & 18,786 & red colobus\_red-capped mangabey & refinement \\
red duiker & \emph{Cephalophus} sp. & genus & $\geqslant$ LC & 633,121 & red duiker &  \\
red-capped mangabey & \emph{Cercocebus torquatus} & species & EN & 10,476 & red colobus\_red-capped mangabey & refinement \\
red-tailed monkey$^\dagger$ & \emph{Cercopithecus ascanius} & species & LC & 619 & monkey & refinement \\
rockfowl & \emph{Picathartes} & genus & $\geqslant$ VU & 1,921 & bird & refinement \\
rodent & \emph{Rodentia} & order & $\geqslant$ LC & 70,027 & rodent &  \\
serval & \emph{Leptailurus serval} & species & LC & 3,664 & serval &  \\
side-striped jackal & \emph{Canis adustus} & species & LC & 6,560 & side-striped jackal &  \\
sitatunga & \emph{Tragelaphus spekii} & species & LC & 2,579 &  & new class \\
spot-nosed monkey & \emph{Cercopithecus nictitans} & species & LC & 1,489 & monkey & refinement \\
spotted hyena & \emph{Crocuta crocuta} & species & LC & 18,234 & spotted hyena &  \\
squirrel & \emph{Sciuridae, Anomaluridae} & family & $\geqslant$ LC & 91,131 & squirrel &  \\
turaco$^\dagger$ & \emph{Musophagidae} & family & $\geqslant$ LC & 198 & bird & refinement \\
warthog & \emph{Phacochoerus} sp. & genus & $\geqslant$ LC & 9,320 & bushpig & refinement \\
water chevrotain & \emph{Hyemoschus aquaticus} & species & LC & 23,677 & water chevrotain &  \\
yellow-backed duiker & \emph{Cephalophus silvicultor} & species & LC & 92,400 & yellow-backed duiker &  \\
\end{longtable}

\clearpage
\section{Per-class validation metrics}\label{app:classmetrics}

\scriptsize
\setlength{\tabcolsep}{3pt}
\renewcommand{\arraystretch}{0.95}

\captionof{table}{DeepForestVisionV2 per-class metrics and support on the validation set.}
\label{tab:val_class_metrics}

\begin{multicols}{2}

\begin{tabularx}{\linewidth}{@{}>{\raggedright\arraybackslash}Xrrrr@{}}
\toprule
\textbf{Class} & \textbf{Precision} & \textbf{Recall} & \textbf{F1} & \textbf{Support} \\
\midrule
red colobus & \num{0.994} & \num{0.945} & \num{0.969} & \num[round-precision=0]{1654} \\
genet & \num{0.956} & \num{0.861} & \num{0.906} & \num[round-precision=0]{1649} \\
gorilla & \num{0.907} & \num{0.667} & \num{0.769} & \num[round-precision=0]{1319} \\
mandrill & \num{0.843} & \num{0.982} & \num{0.907} & \num[round-precision=0]{1158} \\
red-capped mangabey & \num{0.985} & \num{0.874} & \num{0.926} & \num[round-precision=0]{1155} \\
blue duiker & \num{0.824} & \num{0.951} & \num{0.883} & \num[round-precision=0]{1141} \\
spotted hyena & \num{0.881} & \num{0.882} & \num{0.881} & \num[round-precision=0]{1120} \\
guineafowl & \num{0.782} & \num{0.936} & \num{0.852} & \num[round-precision=0]{1117} \\
rodent & \num{0.837} & \num{0.909} & \num{0.872} & \num[round-precision=0]{1112} \\
baboon & \num{0.847} & \num{0.870} & \num{0.858} & \num[round-precision=0]{1091} \\
water chevrotain & \num{0.917} & \num{0.936} & \num{0.927} & \num[round-precision=0]{1085} \\
porcupine & \num{0.885} & \num{0.918} & \num{0.901} & \num[round-precision=0]{1076} \\
bushpig & \num{0.846} & \num{0.945} & \num{0.893} & \num[round-precision=0]{1060} \\
goat & \num{0.863} & \num{0.985} & \num{0.920} & \num[round-precision=0]{1045} \\
black guineafowl & \num{0.810} & \num{0.894} & \num{0.850} & \num[round-precision=0]{1013} \\
red duiker & \num{0.764} & \num{0.973} & \num{0.856} & \num[round-precision=0]{1002} \\
african elephant & \num{0.864} & \num{0.877} & \num{0.871} & \num[round-precision=0]{993} \\
bushbuck & \num{0.924} & \num{0.848} & \num{0.884} & \num[round-precision=0]{987} \\
francolin & \num{0.906} & \num{0.809} & \num{0.855} & \num[round-precision=0]{986} \\
mongoose & \num{0.726} & \num{0.815} & \num{0.768} & \num[round-precision=0]{970} \\
agile mangabey & \num{0.892} & \num{0.844} & \num{0.867} & \num[round-precision=0]{954} \\
squirrel & \num{0.749} & \num{0.948} & \num{0.837} & \num[round-precision=0]{949} \\
cattle & \num{0.958} & \num{0.854} & \num{0.903} & \num[round-precision=0]{934} \\
lhoests monkey & \num{0.861} & \num{0.872} & \num{0.866} & \num[round-precision=0]{927} \\
african buffalo & \num{0.760} & \num{0.922} & \num{0.833} & \num[round-precision=0]{888} \\
yellow-backed duiker & \num{0.778} & \num{0.910} & \num{0.839} & \num[round-precision=0]{876} \\
chimpanzee & \num{0.668} & \num{0.841} & \num{0.744} & \num[round-precision=0]{824} \\
hyrax & \num{0.972} & \num{0.863} & \num{0.914} & \num[round-precision=0]{801} \\
black-and-white colobus & \num{0.993} & \num{0.841} & \num{0.911} & \num[round-precision=0]{793} \\
african golden cat & \num{0.934} & \num{0.706} & \num{0.804} & \num[round-precision=0]{707} \\
\bottomrule
\end{tabularx}

\columnbreak

\begin{tabularx}{\linewidth}{@{}>{\raggedright\arraybackslash}Xrrrr@{}}
\toprule
\textbf{Class} & \textbf{Precision} & \textbf{Recall} & \textbf{F1} & \textbf{Support} \\
\midrule
hippopotamus & \num{0.926} & \num{0.834} & \num{0.878} & \num[round-precision=0]{706} \\
warthog & \num{0.982} & \num{0.882} & \num{0.929} & \num[round-precision=0]{676} \\
side-striped jackal & \num{0.975} & \num{0.623} & \num{0.760} & \num[round-precision=0]{676} \\
black-legged mongoose & \num{0.837} & \num{0.816} & \num{0.826} & \num[round-precision=0]{548} \\
dog & \num{0.818} & \num{0.708} & \num{0.759} & \num[round-precision=0]{514} \\
leopard & \num{0.923} & \num{0.907} & \num{0.915} & \num[round-precision=0]{492} \\
dove & \num{0.915} & \num{0.771} & \num{0.837} & \num[round-precision=0]{432} \\
passeriform & \num{0.910} & \num{0.609} & \num{0.730} & \num[round-precision=0]{417} \\
pangolin & \num{0.777} & \num{0.651} & \num{0.708} & \num[round-precision=0]{401} \\
nkulengu rail & \num{0.763} & \num{0.845} & \num{0.802} & \num[round-precision=0]{388} \\
pelecaniform & \num{0.989} & \num{0.930} & \num{0.958} & \num[round-precision=0]{372} \\
serval & \num{0.819} & \num{0.833} & \num{0.826} & \num[round-precision=0]{354} \\
blue monkey & \num{0.782} & \num{0.726} & \num{0.753} & \num[round-precision=0]{351} \\
grey-cheeked mangabey & \num{0.794} & \num{0.781} & \num{0.787} & \num[round-precision=0]{306} \\
sitatunga & \num{0.904} & \num{0.548} & \num{0.683} & \num[round-precision=0]{259} \\
bird of prey & \num{0.885} & \num{0.889} & \num{0.887} & \num[round-precision=0]{199} \\
civet & \num{0.815} & \num{0.928} & \num{0.868} & \num[round-precision=0]{195} \\
moustached monkey & \num{0.840} & \num{0.757} & \num{0.797} & \num[round-precision=0]{181} \\
honey badger & \num{0.578} & \num{0.325} & \num{0.416} & \num[round-precision=0]{160} \\
casqued hornbill & \num{0.961} & \num{0.892} & \num{0.925} & \num[round-precision=0]{139} \\
rockfowl & \num{0.757} & \num{0.879} & \num{0.813} & \num[round-precision=0]{124} \\
aardvark & \num{0.974} & \num{0.701} & \num{0.815} & \num[round-precision=0]{107} \\
otter & \num{0.893} & \num{0.684} & \num{0.775} & \num[round-precision=0]{98} \\
spot-nosed monkey & \num{0.515} & \num{0.850} & \num{0.642} & \num[round-precision=0]{80} \\
red-tailed monkey & \num{0.636} & \num{0.538} & \num{0.583} & \num[round-precision=0]{78} \\
galago & \num{0.810} & \num{0.523} & \num{0.636} & \num[round-precision=0]{65} \\
hornbill & \num{0.548} & \num{0.727} & \num{0.625} & \num[round-precision=0]{55} \\
duck & \num{1.000} & \num{0.710} & \num{0.830} & \num[round-precision=0]{31} \\
turaco & \num{0.857} & \num{0.571} & \num{0.686} & \num[round-precision=0]{21} \\
potto & \num{0.400} & \num{0.625} & \num{0.488} & \num[round-precision=0]{16} \\
crane & \num{1.000} & \num{1.000} & \num{1.000} & \num[round-precision=0]{5} \\
\bottomrule
\end{tabularx}

\end{multicols}
\end{landscape}

%% file: DFV2.bib
@article{ahumada2020wildlife,
  author  = {Ahumada, Jorge A. and Fegraus, Eric and Birch, Tanya and Flores, Nicole and Kays, Roland and O'Brien, Timothy G. and Palmer, Jonathan and Schuttler, Stephanie and Zhao, Jennifer Y. and Jetz, Walter and others},
  title   = {Wildlife Insights: A Platform to Maximize the Potential of Camera Trap and Other Passive Sensor Wildlife Data for the Planet},
  journal = {Environmental Conservation},
  year    = {2020},
  volume  = {47},
  number  = {1},
  pages   = {1--6},
  doi     = {10.1017/S0376892919000298}
}

@article{beery2019pipeline,
  author  = {Beery, Sara and Morris, Dan and Yang, Siyu},
  title   = {Efficient Pipeline for Camera Trap Image Review},
  journal = {arXiv},
  year    = {2019},
  eprint  = {1907.06772},
  archivePrefix = {arXiv},
  primaryClass  = {cs.CV}
}

@article{dorne2025zamba,
  author  = {Dorne, Emily and others},
  title   = {Zamba: Computer vision for wildlife conservation},
  journal = {Proceedings of the Python in Science Conference (SciPy)},
  year    = {2025},
  howpublished = {\url{https://proceedings.scipy.org/articles/crcw9835}},
  note    = {Accessed 2026-01-04}
}

@article{whytock2021robust,
  author  = {Whytock, Robin C. and {\'S}wie{\.z}ewski, J{\k{e}}drzej and Zwerts, Joeri A. and others},
  title   = {Robust ecological analysis of camera trap data labelled by a machine learning model},
  journal = {Methods in Ecology and Evolution},
  year    = {2021},
  volume  = {12},
  pages   = {1080--1092},
  doi     = {10.1111/2041-210X.13576}
}

@article{whytock2023realtime,
  author  = {Whytock, Robin C. and Suijten, Thijs and van Deursen, Tim and {\'S}wie{\.z}ewski, J{\k{e}}drzej and others},
  title   = {Real-time alerts from AI-enabled camera traps using the Iridium satellite network: A case-study in Gabon, Central Africa},
  journal = {Methods in Ecology and Evolution},
  year    = {2023},
  volume  = {14},
  number  = {3},
  pages   = {867--874},
  doi     = {10.1111/2041-210X.14036}
}

@article{mckaughan2025cropprimates,
  author  = {McKaughan, J. E. T. and Stephens, P. A. and Hill, R. A.},
  title   = {Estimating Abundance of Crop-Foraging Primates in Anthropogenic Landscapes Using Camera Traps},
  journal = {American Journal of Primatology},
  year    = {2025},
  volume  = {87},
  pages   = {e70087},
  doi     = {10.1002/ajp.70087}
}

@article{obrien2008birds,
  author  = {O'Brien, Timothy G. and Kinnaird, Margaret F.},
  title   = {A picture is worth a thousand words: the application of camera trapping to the study of birds},
  journal = {Bird Conservation International},
  year    = {2008},
  volume = {18(S1)},
    doi   = {10.1017/S0959270908000348}
}

@article{SpeciesNet,
author = {Gadot, Tomer and Istrate, S and Kim, Hyungwon and Morris, Dan and Beery, Sara and Birch, Tanya and Ahumada, Jorge},
title = {To crop or not to crop: Comparing whole-image and cropped classification on a large dataset of camera trap images},
journal = {IET Computer Vision},
volume = {18},
number = {8},
pages = {1193-1208},
doi = {https://doi.org/10.1049/cvi2.12318},
url = {https://ietresearch.onlinelibrary.wiley.com/doi/abs/10.1049/cvi2.12318},
eprint = {https://ietresearch.onlinelibrary.wiley.com/doi/pdf/10.1049/cvi2.12318},
year = {2024}
}

@article{DFV,
author = {Magaldi, Hugo and Cornette, Raphaël and Tibesigwa, John Justice and Katumba, Raymond and Rugonge, Harold and Amarasekaran, Bala and Anderson, Naomi and Cappelle, Noémie and Cardoso, Anabelle W. and Cornélis, Daniel and Deschner, Tobias and Fonteyn, Davy and Garriga, Rosa M. and van Lunteren, Peter and Rufray, Xavier and Vanthomme, Hadrien and Zwerts, Joeri A. and Krief, Sabrina},
title = {DeepForestVision: Automated wildlife identification for camera traps of African tropical forests},
journal = {Ecological Solutions and Evidence},
volume = {6},
number = {4},
pages = {e70167},
doi = {https://doi.org/10.1002/2688-8319.70167},
url = {https://besjournals.onlinelibrary.wiley.com/doi/abs/10.1002/2688-8319.70167},
eprint = {https://besjournals.onlinelibrary.wiley.com/doi/pdf/10.1002/2688-8319.70167},
note = {e70167 ESO-25-07-156.R1},
year = {2025}
}

@article{PlatformsEvaluation,
author = {Vélez, Juliana and McShea, William and Shamon, Hila and Castiblanco-Camacho, Paula J. and Tabak, Michael A. and Chalmers, Carl and Fergus, Paul and Fieberg, John},
title = {An evaluation of platforms for processing camera-trap data using artificial intelligence},
journal = {Methods in Ecology and Evolution},
volume = {14},
number = {2},
pages = {459-477},
doi = {https://doi.org/10.1111/2041-210X.14044},
url = {https://besjournals.onlinelibrary.wiley.com/doi/abs/10.1111/2041-210X.14044},
eprint = {https://besjournals.onlinelibrary.wiley.com/doi/pdf/10.1111/2041-210X.14044},
year = {2023}
}

@article{vanLunteren2023, doi = {10.21105/joss.05581}, url = {https://doi.org/10.21105/joss.05581}, year = {2023}, publisher = {The Open Journal}, volume = {8}, number = {88}, pages = {5581}, author = {van Lunteren, Peter}, title = {AddaxAI: A no-code platform to train and deploy custom YOLOv5 object detection models}, journal = {Journal of Open Source Software} }

@misc{simeoni2025dinov3,
      title={DINOv3}, 
      author={Oriane Siméoni and Huy V. Vo and Maximilian Seitzer and Federico Baldassarre and Maxime Oquab and Cijo Jose and Vasil Khalidov and Marc Szafraniec and Seungeun Yi and Michaël Ramamonjisoa and Francisco Massa and Daniel Haziza and Luca Wehrstedt and Jianyuan Wang and Timothée Darcet and Théo Moutakanni and Leonel Sentana and Claire Roberts and Andrea Vedaldi and Jamie Tolan and John Brandt and Camille Couprie and Julien Mairal and Hervé Jégou and Patrick Labatut and Piotr Bojanowski},
      year={2025},
      eprint={2508.10104},
      archivePrefix={arXiv},
      primaryClass={cs.CV},
      url={https://arxiv.org/abs/2508.10104}, 
}

@book{CIRAD,
  author    = {Cornelis, Daniel},
  title     = {Camera trap survey data of the project ``Sustainable Wildlife Management'' in Gabon (Mulundu Department)},
  publisher = {CIRAD Dataverse},
  year      = {2023},
  doi       = {10.18167/DVN1/TNBXCN}
}

@article{zwerts_fsc-certified_2024,
  author  = {Zwerts, Joeri A. and Sterck, E. H. M. and Verweij, Pita A. and Maisels, Fiona and van der Waarde, Jaap and Geelen, Emma A. M. and Tchoumba, Georges Belmond and Donfouet Zebaze, Hermann Frankie and van Kuijk, Marijke},
  title   = {{FSC}-certified forest management benefits large mammals compared to non-{FSC}},
  journal = {Nature},
  year    = {2024},
  volume  = {628},
  number  = {8008},
  pages   = {563--568},
  doi     = {10.1038/s41586-024-07257-8}
}

@article{suss_hatlauf_2024,
  author    = {Suss, L. and Hatlauf, J.},
  title     = {Focus on carnivore communities: photo traps and data analysis in biodiversity research},
  journal   = {Acta Zoobot},
  year      = {2024},
  volume    = {160}
}

@article{hatlauf_hacklaender_2016,
  author    = {Hatlauf, J. and Hackländer, K.},
  title     = {Preliminary results of golden jackal (Canis aureus) survey in Austria},
  journal   = {Beiträge zur Jagd- und Wildforschung},
  year      = {2016},
  volume    = {41},
  pages     = {295--306}
}

@article{tropical_forests_monitoring,
    doi = {10.1371/journal.pone.0073707},
    author = {Ahumada, Jorge A. AND Hurtado, Johanna AND Lizcano, Diego},
    journal = {PLOS ONE},
    publisher = {Public Library of Science},
    title = {Monitoring the Status and Trends of Tropical Forest Terrestrial Vertebrate Communities from Camera Trap Data: A Tool for Conservation},
    year = {2013},
    month = {09},
    volume = {8},
    url = {https://doi.org/10.1371/journal.pone.0073707},
    pages = {1-10},
    number = {9},

}

@article{tuia_perspectives_2022,
  author  = {Tuia, Devis and Kellenberger, Benjamin and Beery, Sara and Costelloe, Blair R. and Zuffi, Silvia and Risse, Benjamin and Mathis, Alexander and Mathis, Mackenzie W. and van Langevelde, Frank and Burghardt, Tilo and Kays, Roland and Klinck, Holger and Wikelski, Martin and Couzin, Iain D. and van Horn, Grant and Crofoot, Margaret C. and Stewart, Charles V. and Berger-Wolf, Tanya},
  title   = {Perspectives in machine learning for wildlife conservation},
  journal = {Nature Communications},
  year    = {2022},
  volume  = {13},
  number  = {1},
  pages   = {792},
  doi     = {10.1038/s41467-022-27980-y}
}

@article{walton_camera_2022,
  author  = {Walton, Ben J. and Findlay, Leah J. and Hill, Russell A.},
  title   = {Camera traps and guard observations as an alternative to researcher observation for studying anthropogenic foraging},
  journal = {Ecology and Evolution},
  year    = {2022},
  volume  = {12},
  number  = {4},
  pages   = {e8808},
  doi     = {10.1002/ece3.8808}
}

@article{Moore_arboreal,
author = {Moore, J. F. and Pine, W. E. and Mulindahabi, F. and Niyigaba, P. and Gatorano, G. and Masozera, M. K. and Beaudrot, L.},
title = {Comparison of species richness and detection between line transects, ground camera traps, and arboreal camera traps},
journal = {Animal Conservation},
volume = {23},
number = {5},
pages = {561-572},
doi = {https://doi.org/10.1111/acv.12569},
url = {https://zslpublications.onlinelibrary.wiley.com/doi/abs/10.1111/acv.12569},
eprint = {https://zslpublications.onlinelibrary.wiley.com/doi/pdf/10.1111/acv.12569},
year = {2020}
}

@article{pebsworth_selecting_2021,
  author  = {Pebsworth, Paula A. and Gruber, Thibaud and Miller, Joshua D. and Zuberb"uhler, Klaus and Young, Sera L.},
  title   = {Selecting between iron-rich and clay-rich soils: a geophagy field experiment with black-and-white colobus monkeys in the Budongo Forest Reserve, Uganda},
  journal = {Primates},
  year    = {2021},
  volume  = {62},
  number  = {1},
  pages   = {133--142},
  doi     = {10.1007/s10329-020-00845-y}
}

@inproceedings{deng2009imagenet,
  title     = {ImageNet: A Large-Scale Hierarchical Image Database},
  author    = {Deng, Jia and Dong, Wei and Socher, Richard and Li, Li-Jia and Li, Kai and Fei-Fei, Li},
  booktitle = {Proceedings of the IEEE Conference on Computer Vision and Pattern Recognition (CVPR)},
  pages     = {248--255},
  year      = {2009},
  doi       = {10.1109/CVPR.2009.5206848}
}
